\documentclass[letterpaper, journal, twoside]{IEEEtran}

\IEEEoverridecommandlockouts                              

\usepackage{subcaption}
\usepackage{hyperref}       
\usepackage{booktabs}       
\usepackage{placeins}
\usepackage{graphicx}
\usepackage[acronym,shortcuts]{glossaries}
\usepackage{comment}
\usepackage{enumerate}
\usepackage[capitalise]{cleveref}
\usepackage[table,xcdraw,dvipsnames]{xcolor}
\usepackage{siunitx}
\usepackage{bm} 
\usepackage{amssymb} 
\usepackage{amsmath}
\usepackage[english]{babel}
\usepackage{multirow}
\usepackage{pifont}
\usepackage{algorithm}
\usepackage{algorithmic}
\usepackage{cuted}
\usepackage{cite}
\usepackage[normalem]{ulem}

\DeclareMathOperator*{\argmin}{arg\,min}
\definecolor{lightgray}{gray}{0.9}

\def\BibTeX{{\rm B\kern-.05em{\sc i\kern-.025em b}\kern-.08em
    T\kern-.1667em\lower.7ex\hbox{E}\kern-.125emX}}
\newcommand{\todo}[2][TODO: ]{\textcolor{orange}{#1 #2}}
\newcommand{\rev}[1]{\textcolor{black}{#1}}
\usepackage[normalem]{ulem} 
\newcommand{\revdel}[1]{\textcolor{blue}{\sout{#1}}}%

\renewcommand{\revdel}[1]{}

\usepackage{eso-pic}

\newcommand\AtPageUpperCenterNotice[1]{%
  \AtPageUpperLeft{%
    \put(\LenToUnit{0.5\paperwidth},\LenToUnit{-2cm}){\makebox[0pt]{#1}}%
  }%
}

\AddToShipoutPictureBG*{%
  \AtPageUpperCenterNotice{%
    \parbox[b][2cm][c]{\paperwidth}{%
      \centering
      \fontsize{12}{14}\selectfont
      \color{gray!50}
      This paper has been accepted for publication in\\
      IEEE Robotics and Automation Letters (RA-L) \copyright{}IEEE.
    }%
  }%
}

\AddToShipoutPictureBG*{
  \AtPageLowerLeft{%
    \raisebox{25pt}{\makebox[\paperwidth]{\begin{minipage}{21cm}\centering
    \fontsize{10}{8}\selectfont
    \textcolor{gray!50}{%
      \copyright{} 2025 IEEE. Personal use of this material is permitted.
      Permission from IEEE must be obtained for all other uses, in any current or future media,
      including reprinting/republishing this material for advertising or promotional purposes,
      creating new collective works, for resale or redistribution to servers or lists,
      or reuse of any copyrighted component of this work in other works.
    }
    \end{minipage}}}%
  }
}
\title{\LARGE \bf \vspace{6mm}
Planar Velocity Estimation for Fast-Moving Mobile Robots Using Event-Based Optical Flow}

\author{Liam Boyle\IEEEauthorrefmark{1}\IEEEauthorrefmark{2}, Jonas Kühne\IEEEauthorrefmark{1}\IEEEauthorrefmark{2}, Nicolas Baumann\IEEEauthorrefmark{1}\IEEEauthorrefmark{2}, Niklas Bastuck\IEEEauthorrefmark{1}, and Michele Magno\IEEEauthorrefmark{1} \vspace{-8mm}
\thanks{Manuscript received: February, 14$^{th}$, 2025; Revised April, 18$^{th}$, 2025; Accepted May, 15$^{th}$, 2025. This paper was recommended for publication by Editor Pascal Vasseur upon evaluation of the Associate Editor and Reviewers' comments.}
\thanks{\IEEEauthorrefmark{1}Liam Boyle, Jonas Kühne, Nicolas Baumann, Niklas Bastuck, and Michele Magno are associated with the Center for Project-Based Learning, D-ITET, ETH Zurich.
}
\thanks{\emph{\IEEEauthorrefmark{2}Contributed equally to this work. (Corresponding author: Liam Boyle.)}}
}

\begin{document}
\newacronym{lidar}{LiDAR}{Light Detection and Ranging}
\newacronym{radar}{RADAR}{Radio Detection and Ranging}
\newacronym{mpc}{MPC}{Model Predictive Control}
\newacronym{mpcc}{MPCC}{Model Predictive Contouring Control}
\newacronym{iot}{IoT}{Internet of Things}
\newacronym{bev}{BEV}{Bird's-Eye View}
\newacronym{sota}{SotA}{State-of-the-Art}
\newacronym{gpu}{GPU}{Graphics Processing Unit}
\newacronym{fps}{FPS}{Frames Per Second}
\newacronym{kf}{KF}{Kalman Filter}
\newacronym{ads}{ADS}{Autonomous Driving Systems}
\newacronym{iac}{IAC}{Indy Autonomous Challenge}
\newacronym{fsd}{FSD}{Formula Student Driverless}
\newacronym{rrt*}{RRT*}{Rapidly exploring Random Tree Star}
\newacronym{mgbt}{MGBT}{Multilayer Graph-Based Trajectory}
\newacronym{ftg}{FTG}{Follow The Gap}
\newacronym{gbo}{GBO}{Graph-Based Overtake}
\newacronym{gp}{GP}{Gaussian Process}
\newacronym{rbf}{RBF}{Radial Basis Function}
\newacronym{map}{MAP}{Model- and Acceleration-based Pursuit}
\newacronym{sqp}{SQP}{Sequential Quadratic Programming}
\newacronym{roc}{RoC}{Region of Collision}
\newacronym{adas}{ADAS}{Advanced Driving Assistance Systems}
\newacronym{rpm}{RPM}{Revolutions Per Minute}
\newacronym{erpm}{ERPM}{Electric Revolutions Per Minute}
\newacronym{imu}{IMU}{Inertial Measurement Unit}
\newacronym{ekf}{EKF}{Extended Kalman Filter}
\newacronym{gps}{GPS}{Global Positioning System}
\newacronym{gnss}{GNSS}{Global Navigation Satellite Systems}
\newacronym{rtk}{RTK}{Real-Time Kinematic positioning}
\newacronym{lio}{LIO}{LiDAR Inertial Odometry}
\newacronym{vio}{VIO}{Visual Inertial Odometry}
\newacronym{evio}{eVIO}{event Visual Inertial Odometry}
\newacronym{eof}{eOF}{event Optical Flow}
\newacronym{of}{OF}{Optical Flow}
\newacronym{uav}{UAV}{Unmanned Aerial Vehicle}
\newacronym{fov}{FoV}{Field of View}
\newacronym{ir}{IR}{Infra Red}
\newacronym{ros}{ROS}{Robot Operating System}
\newacronym{cpu}{CPU}{Central Processing Unit}
\newacronym{ml}{ML}{Machine Learning}
\newacronym{cots}{CotS}{Commercial off-the-Shelf}
\newacronym{ransac}{RANSAC}{Random Sample Consensus}
\newacronym{rmse}{RMSE}{Root Mean Squared Error}
\newacronym{uslam}{USLAM}{Ultimate SLAM}
\newacronym{epe}{EPE}{End-Point Error}
\newacronym{ae}{AE}{Angular Error}
\newacronym{obc}{OBC}{On-Board Computer}
\newacronym{tof}{ToF}{Time of Flight}


\markboth{IEEE Robotics and Automation Letters. Preprint Version. Accepted May, 2025}
{Boyle \MakeLowercase{\textit{et al.}}: Planar Velocity Estimation Using Event Optical Flow}
\maketitle


\thispagestyle{empty}
\pagestyle{empty}

\begin{abstract}
Accurate velocity estimation is critical in mobile robotics, particularly for driver assistance systems and autonomous driving. Wheel odometry fused with Inertial Measurement Unit (IMU) data is a widely used method for velocity estimation, however, it typically requires strong assumptions, such as non-slip steering, or complex vehicle dynamics models that do not hold under varying environmental conditions, like slippery surfaces. We introduce an approach to velocity estimation that is decoupled from wheel-to-surface traction assumptions by leveraging planar kinematics in combination with optical flow from event cameras pointed perpendicularly at the ground. The asynchronous \(\mu\)-second latency and high dynamic range of event cameras make them highly robust to motion blur, a common challenge in vision-based perception techniques for autonomous driving. The proposed method is evaluated through in-field experiments on a 1:10 scale autonomous racing platform and compared to precise motion capture data \rev{demonstrating not only performance on par with the State-of-the-Art Event-VIO method but also a $38.3\%$ improvement in lateral error.}  \revdel{which demonstrates a $6.5\%$ improvement in longitudinal and a $38.3\%$ improvement in lateral error over the State-of-the-Art Event-VIO method.} Qualitative experiments at highway speeds of up to \SI{32}{\metre\per\second} further confirm the effectiveness of our approach, indicating significant potential for real-world deployment.

\end{abstract}

\setlength{\tabcolsep}{3pt}

\section{Introduction}
\IEEEPARstart{I}n automotive applications and mobile robotics, precise velocity estimation plays a crucial role in optimizing the functionality and safety of \gls{adas} and \gls{ads} \cite{battistini2022enhancing}. Although cars universally employ speedometer gauges, suggesting that the velocity estimation task is solved, these methods are often imprecise and lack robustness against varying environmental conditions, such as slippery surfaces. Typically, velocity data on wheeled robots is derived from wheel odometry, which measures wheel rotation speed (e.g., wheel encoders) or motor \gls{rpm}. However, these techniques rely on factors such as the wheel circumference and can fail on slippery surfaces such as dusty or icy roads due to their reliance on the assumption of ideal wheel-to-surface traction \cite{gp_wheel_slip, ground_slam}. Wheel-encoder-based velocity measurements are limited to longitudinal velocity, as they rely on counting wheel rotations, which do not capture lateral motion.

\begin{figure}[t]
    \centering
    \includegraphics[width=\linewidth]{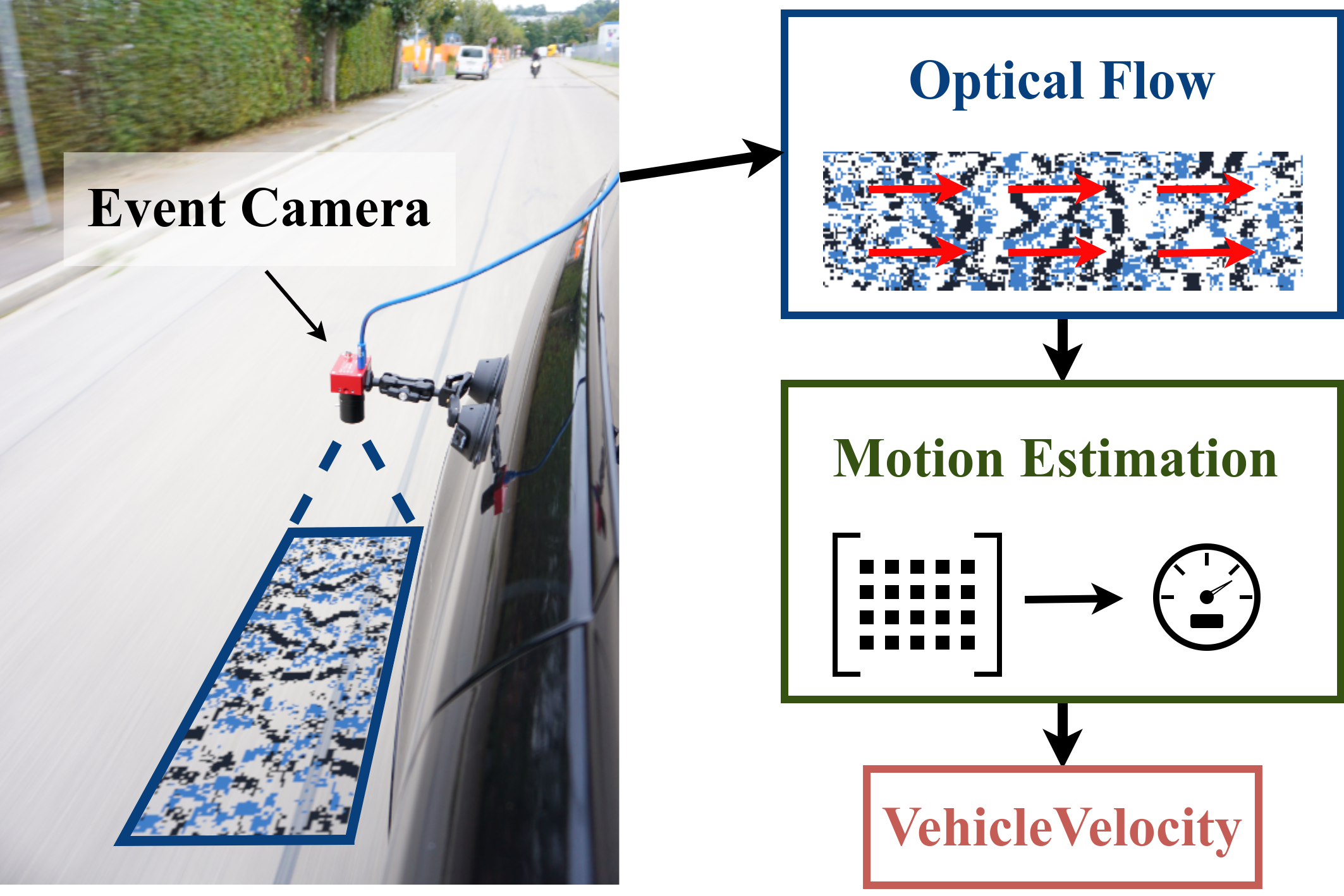}
    \caption{Overview of the proposed method. Depicted on the left is the experimental setup used during the full-scale highway experiment as well as an example event frame captured at \SI{32}{\metre\per\second} and accumulated over \SI{100}{\micro\second}.}
    \label{fig:title-image}
\end{figure}

To enhance robustness of velocity estimation with respect to longitudinal and lateral wheel-slip, some approaches integrate \gls{imu} data with wheel odometry in an \gls{ekf} with an appropriate motion model \cite{robot_localizer, forzaeth}. Nonetheless, these solutions are limited by \gls{imu} sensor noise and drift, requiring strong underlying assumptions, such as non-slip \textit{Ackermann} steering \cite{ciot}. Although these assumptions may be valid in low-speed scenarios, where lateral velocity can be considered negligible, they generally don't hold at high speeds \cite{indy_3d_vel}. In such instances, accurate lateral velocity estimation becomes essential for enhancing vehicle performance through advanced control strategies, such as \gls{mpc} \cite{mpcc}, or for enabling slip detection in \gls{adas} safety systems \cite{battistini2022enhancing}.


To enable lateral velocity estimation, complex motion models incorporating the notoriously difficult-to-model tire dynamics \cite{story_of_modelmismatch} would have to be employed. This approach is often intractable outside of controlled environments due to the environmentally variable surface slip, which would significantly alter the motion model parameters \cite{indy_3d_vel}. Hence, the only practical way to derive accurate longitudinal and lateral velocity estimations is to deduce the odometry from velocity measurements that are decoupled from the wheels. Technologies such as \gls{gps} could theoretically enable this, but \gls{gnss} are typically too imprecise and offer low update rates --- even with enhancements like \gls{rtk} \cite{li2022real} --- as well as unreliable \gls{gnss} coverage (e.g. in a tunnel), to meet the demanding requirements of \gls{ads} and \gls{adas}. As alternatives, \gls{lio} \cite{kissicp} and \gls{vio} \cite{uslam, orbslam} methods, which rely on \gls{lidar} and camera sensors, respectively, appear promising. These techniques primarily estimate the robot's pose through scan-matching algorithms in \gls{lio} \cite{kissicp}, or by matching features from image key points in \gls{vio} \cite{uslam, orbslam}. However, \gls{lio} methods are known to work poorly in degenerative environments \cite{lio_icra_degenerate}, such as tunnels or large open spaces, which regularly occur in driving scenarios.

Event cameras are a novel type of sensor that captures changes in intensity rather than absolute intensity values. Unlike traditional frame-based cameras, they offer continuous, asynchronous data output, making them well-suited for dynamic environments. The characteristics of event cameras, such as their high dynamic range and low latency, have shown great potential for automotive applications \cite{event_automotive_nature, dsec}. While currently niche, event cameras are expected to become significantly more affordable once mass production begins (in the price range of traditional cameras), making them a promising alternative to traditional vision sensors, given their characteristics \cite{event_cheap}. Consequently, the event camera sensor modality will be leveraged in velocity estimation techniques throughout this work.

We present an accurate velocity estimation algorithm based on \gls{eof}, utilizing an event camera pointed directly at the road surface and mounted close to the ground. This sensor placement allows the perceived scenery to be interpreted as a single rigid body, drastically simplifying the velocity estimation task. The planar assumption is well-justified, as road cars always move in a plane parallel to the road surface. While the close proximity to the ground can cause rapid motions the event sensor modality represents an ideal sensor to capture such fast-moving scenery \cite{event_automotive_nature, emergent_technologies, dsec}. 
We demonstrate through quantitative experiments that our approach exceeds the performance of the \gls{sota} monocular \gls{evio} method, especially in lateral velocity estimation, while not suffering from the same drawbacks of sensor initialization \cite{gps_imu_init}. The main contributions of this work are:


\begin{enumerate}[I]
    \item \textbf{Event-based Approach to Velocity Estimation:}
    We frame the problem of velocity estimation as a case of planar kinematics and propose a novel and simple approach to estimate a vehicle's velocity using \gls{eof} measurements of a camera capturing the road surface.
    \item \textbf{Comparison of Velocity Estimation Methods:}
    We compare our newly proposed method against the \gls{sota} \gls{evio} method both qualitatively and quantitatively. Our method achieves an improvement of $6.5\%$ \gls{rmse} in longitudinal and $38.3\%$ \gls{rmse} in lateral direction, proving the efficacy of our approach.
    \item \textbf{Qualitative Evaluation at Highway Speeds:}
    Real-world experiments on a public highway, driving with up to \SI{32}{\metre\per\second}, were conducted to showcase our method's estimation accuracy, even at high speeds.
    \item \textbf{Open-Source Implementation:} Link will be added upon acceptance.
\end{enumerate}

\section{Related Work} \label{sec:rw}

\subsection{VIO-Based Velocity Estimation} \label{sec:rw_vio}
The \gls{uav} research domain has significantly advanced \gls{vio} research for operations in environments where \gls{gnss} is unavailable, as drones depend heavily on precise positional data \cite{delmerico2018benchmark}. Given that wheel odometry is not applicable for \gls{uav}s, \gls{vio} provides a viable alternative through its combined use of exteroceptive and proprioceptive sensing modalities \cite{delmerico2018benchmark, orbslam}.

In pure Visual Odometry, the displacement of a pixel or a feature from one image frame to the next can be determined either by detecting and matching features like in PTAM \cite{klein2007parallel} or by directly minimizing the photometric error across pixel patches, assuming constant pixel illumination like in DTAM \cite{newcombe2011dtam}. Building on these principles, \gls{vio} estimates the movement of a camera between frames, augmenting the measurements with additional metric acceleration and rotation data through the use of an \gls{imu}. Methods that focus on minimizing the photometric error are known as direct methods, while those that rely on feature detection and matching are referred to as indirect methods \cite{klein2007parallel,newcombe2011dtam}. VINS-Mono is a prominent example of a recent indirect \gls{vio} algorithm \cite{qin2018vins}. Modern algorithms like the ORB-SLAM series use a combination of both concepts, performing fast direct optimization on a frame-to-frame basis and slower feature-based matching and optimization on selected keyframes for more robust and accurate pose estimation \cite{orbslam}. 

\subsection{\rev{Event VIO Algorithms}}

\emph{Ultimate SLAM} \cite{uslam}, ESVIO \cite{chen2023esvio}, PL-EVIO \cite{guan2023pl}, \rev{and Elamin et al. \cite{elamin2024event}} fuse measurements from a conventional camera and \gls{imu} with event camera data to form an \gls{evio} system, allowing them to make accurate velocity estimates, even at very high speeds. 
\revdel{In this work, we use a monocular camera setup. While \emph{Ultimate SLAM} and PL-EVIO provide \gls{sota} \gls{evio} performance, only the code base of \emph{Ultimate SLAM} is openly available.}
\rev{REVIO \cite{wang2022revio} replaces a conventional camera with a range sensor. PL‑EVIO \cite{guan2023pl} augments point features with line‑based event features to exploit the geometry of man‑made structures. Using a monocular setup, we omit stereo methods like ESVIO \cite{chen2023esvio}; although newer \gls{evio} approaches \cite{guan2023pl, elamin2024event} surpass \emph{Utilame SLAM} \cite{uslam}, their code is not publicly available.} Therefore, we consider \emph{Ultimate SLAM} \cite{uslam}, the relevant \gls{sota} baseline for this work.




\subsection{OF-Based Velocity Measurement} \label{sec:rw_of}
\gls{of}-based velocity estimation is prevalent in applications like computer mice, which rely on specialized optical sensors that work best at fixed distances and low speeds (up to \SI{1}{\metre\per\second}) under strong illumination. Due to these limitations, such sensors are rarely used in mobile robotics \cite{dahmen2014odometry,px4flow}. By contrast, the \gls{uav} research domain utilizes \gls{of} to prevent quadcopters from drifting while hovering, exemplified through the \emph{PX4Flow} algorithm and its derivatives \cite{px4flow,kuhne2022parallelizing}. These algorithms are favored due to their simplicity and low complexity compared to \gls{vio}-based methods. \emph{PX4Flow} employs a grey-scale frame camera and a range-finder to calculate velocity based on pixel displacement over sequential images assuming a mostly flat ground, which is effective at the typical operating altitudes of drones. 

\subsection{Event OF Algorithms} \label{sec:rw_eof}
For automotive applications, where realistic speeds of up to \SI{40}{\meter \per \second} (about 140\,km/h) are expected, resulting in a pixel movement of \SI{36950}{pixel \per \second} or \SI{185}{pixels \per frame} (considering a camera at \SI{0.6}{\meter} height with 640 pixels and a \gls{fov} of 60 degrees at \SI{200}{fps} as described in \cref{subsec:g0}), the frame camera technology proves inadequate, as the required exposure time of \SI{0.5}{\milli \second} to \SI{1.0}{\milli \second} (assuming a cloudy day at \SI{400}{Lux}) leads to motion blur \cite{gove2020cmos}. Therefore, we propose employing event-based imaging sensors \cite{li2023emergent} combined with the simplicity and low complexity of \gls{of} techniques for robust and effective velocity estimation in such environments \cite{px4flow,farneback2003two}.

Event-based cameras have been developed and investigated mainly during the last decade. Thus, most approaches to calculating \gls{of} from event data have focused on the use of \gls{ml} techniques \cite{wu2024lightweight,gehrig2021raft}. \Cref{tab:dsec} shows a comparison of recent works on the DSEC-Flow \cite{dsec} benchmark against Farneback \cite{farneback2003two}, a classical optical flow algorithm developed for conventional images. For Farneback to be able to process the asynchronous event data, the accumulation of event data as described in \Cref{sec:eof_implementation} has been applied. While methods like E-RAFT \cite{gehrig2021raft} and IDNet \cite{wu2024lightweight} achieve significant improvements over Farneback their comparatively slow runtime makes them unsuitable for real-time applications. On the other hand, EV-FlowNet \cite{ev_flowNet} represents a more lightweight network but does not achieve significant gains in performance over Farneback.

As it is intended to deploy these methods on a 1:10 scale autonomous racing platform \cite{forzaeth} without any GPU acceleration, the Farneback algorithm appears to strike a good trade-off between latency and accuracy. While \gls{ml}-based approaches like IDNet \cite{wu2024lightweight} and E-RAFT \cite{gehrig2021raft} achieve lower \glspl{epe} (0.84, 1.20) and \glspl{ae} (3.41, 3.75), their high computational demands (120–710 ms latency) require GPU acceleration, making them impractical for resource-limited platforms. In contrast, Farneback achieves a competitive performance (EPE: 2.52, AE: 7.96) with a low CPU latency of 39 ms. Notably, EV-FlowNet \cite{ev_flowNet}, the only \gls{ml}-based event method with a comparable best-case latency (20 ms on GPU), performs on par with Farneback, further validating Farnebacks efficacy for real-time processing.



\begin{table}[t]
\vspace{2mm}
\centering
\resizebox{0.73\columnwidth}{!}{%
\begin{tabular}{l|c|c|cc}
\toprule
\toprule
\multirow{3}{*}{\textbf{Method}}            & \bm{$EPE \downarrow$}                  & \bm{$AE \downarrow$}                  & \multicolumn{2}{c}{\bm{$Latency \downarrow$}}         \\
                                   & \multirow{2}{*}{[\SI{}{pixel}]}              & \multirow{2}{*}{[\SI{}{degrees}]}           & \multicolumn{2}{c}{[\SI{}{\milli\second}]} \\
                                   &                             &                            & CPU & E-GPU\\
\midrule
\midrule
EV-FlowNet \cite{ev_flowNet}       & 2.32                        & 7.90                        & 200 & \textbf{20}\\
IDNet* \cite{wu2024lightweight}    & \textbf{0.84}               & \textbf{3.41}              & 120 & 120\\
E-RAFT* \cite{gehrig2021raft}      & 1.20                         & 3.75                       & 530 & 710\\
Farneback \cite{farneback2003two} & 2.52                        & 7.96                       & \textbf{39}& -\\
\bottomrule
\bottomrule
\end{tabular}%
}
\caption{Evaluation on DSEC-Flow dataset with \gls{epe} (L2
endpoint error in pixels), \gls{ae} (angular error in degrees), and \emph{Latency} (latency in milliseconds) as performance metrics. The listed metrics for EV-FlowNet, IDNet and E-RAFT are reported from \cite{wu2024lightweight} and the * indicates that the configurations with the lowest latency are shown. For Farneback the CPU latency was computed on an \texttt{Intel Core i7-13700k}.}
\label{tab:dsec}
\vspace{-10pt}
\end{table}

\section{Background}\label{subsec:g0}

Event cameras have demonstrated their utility in environmental scene understanding for autonomous driving due to their high dynamic range and low latency in the microsecond range \cite{event_automotive_nature, emergent_technologies, li2023emergent}. This enables them to effectively capture challenging lighting scenarios, such as when entering or exiting tunnels or in low-light conditions, where conventional RGB cameras may be temporarily blinded or fail to function properly. The low latency further facilitates the perception and reaction to highly dynamic obstacles \cite{emergent_technologies, event_automotive_nature}. Hence, the advantages of event cameras for scene understanding in \gls{ads} are well established.

\begin{figure}[t]
    \centering
    \includegraphics[width=0.83\columnwidth]{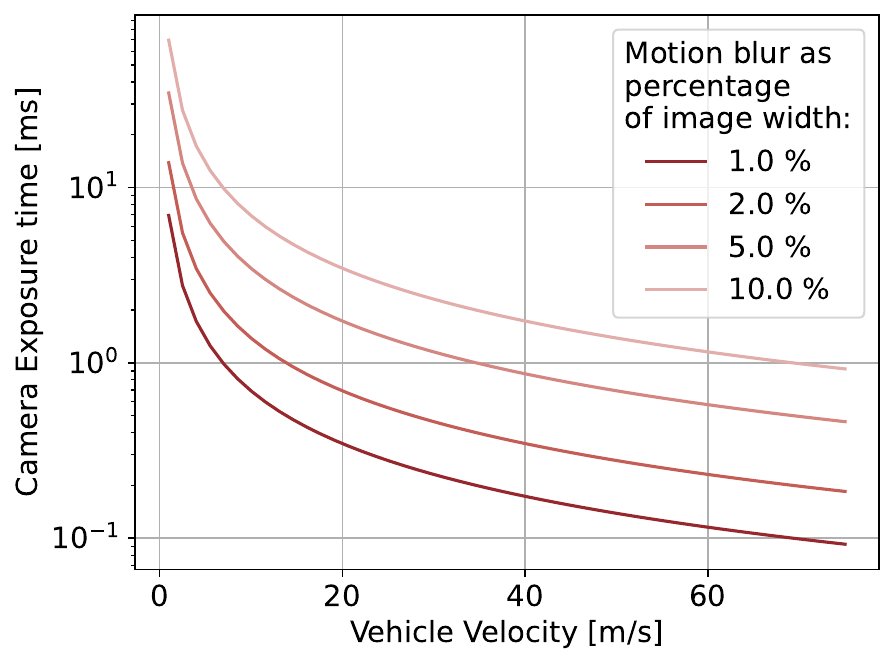}
    \caption{Comparison of the maximum allowable camera exposure time to not exceed the indicated levels of motion blur at different vehicle speeds. A fixed camera height of \SI{0.6}{\metre} and \gls{fov} of \SI{60}{\degree} are assumed.}
    \label{fig:g0}
\end{figure}

However, we emphasize an additional advantage of event cameras for velocity estimation. When pointing a conventional frame camera downwards close to the ground, the observed scene will shift significantly during the camera's exposure time, even when moving at moderate speeds, causing severe motion blur and degrading the quality of downstream \gls{of} calculations. For a given exposure time, the severity of motion blur depends on vehicle speed, camera \gls{fov}, and the height at which the camera is mounted:

\begin{equation}
    \text{Relative Motion Blur} = \frac{t_{exp} \cdot v}{z \cdot 2 \cdot \tan\left(\frac{\alpha}{2}\right)},
    \label{eq:motion_blur}
\end{equation}

where $t_{exp}$ is the exposure time, $z$ is the distance of the camera to the ground, $\alpha$ is the camera's \gls{fov}, and $v$ is the velocity with which the camera is moving.

\begin{figure}[t]
    \centering
    \includegraphics[width=0.49\columnwidth]{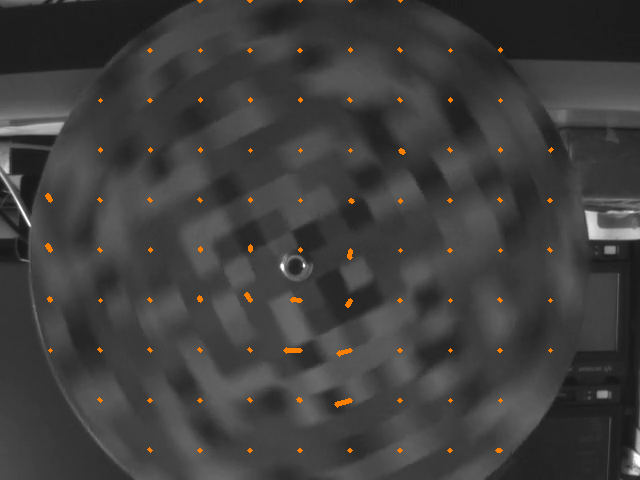}
    \includegraphics[width=0.49\columnwidth]{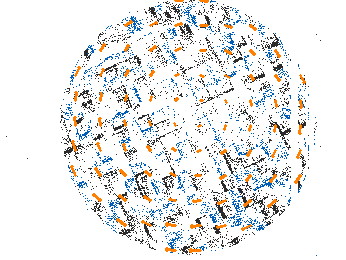}
    \caption{Qualitative \gls{of} visualization of a conventional camera image (left) at 200 \gls{fps} and an accumulated event camera image (right) using an accumulation time of \SI{1}{\milli \second} (\SI{1000}{Hz}). The disk positioned at \SI{0.3}{\meter} from the cameras with a radius of \SI{0.15}{\meter} is spinning at 360.0\,\gls{rpm} for both cameras, resulting in a rotational velocity of \SI{5.65}{\meter \per \second} at the outer edge of the disk. The recordings were taken at \SI{387}{Lux}, with an exposure time of \SI{0.87}{\milli \second} for the conventional camera. \rev{\gls{of} vectors depicted as arrows (orange), pixel-level brightness changes of the event camera visualised in (blue/black).}}
    \label{fig:rgbvsevent}
\end{figure}

\Cref{fig:g0} illustrates the relationship between motion blur, exposure time, and vehicle velocity assuming a fixed camera height and \gls{fov}, set to \SI{0.6}{\meter} and \SI{60}{\degree}, respectively. For an image that is \SI{640}{pixels} wide, a relative motion blur of $1\%$ indicates that image features will be blurred over \SI{6.4}{\rev{pixels}}, which is already a significant amount, as can be seen in Figure \ref{fig:rgbvsevent}. Thus, to reliably measure velocities up to speeds of \SI{40}{\metre\per\second} (approximately highway driving speeds) with less than $1\%$ of relative motion blur requires camera exposure times to be no longer than \SI{170}{\micro\second}. Capturing high-quality images with RGB cameras using such short exposure times requires expensive, specialized cameras and bright light to adequately illuminate the sensor in this short exposure period. Conversely, event cameras, with their high dynamic range and asynchronous data capture, do not suffer from these shortcomings, making them an ideal sensor for this application.

Further, note within \Cref{fig:rgbvsevent} that the conventional camera cannot capture the scene without motion blur such that accurate \gls{of} computation is not possible. According to \Cref{eq:motion_blur} we can calculate the relative motion blur \rev{--- through geometric properties of the scene, including exposure time, camera velocity, depth, and camera field of view ---} to be \SI{1.4}{\%}, hence the pattern is being blurred across 9.1 pixels at the edge of the disk. For the event camera recording, the Farneback-based \gls{eof} algorithm presented in \Cref{sec:eof_implementation} estimates an average angle velocity of \SI{0.03766}{\radian \per frame}, or \SI{37.66}{\radian \per \second} (359.6 \gls{rpm}), which is nearly spot on the actual 360 \gls{rpm}.

\section{Methodology}
This work proposes an accurate \gls{eof}-based approach for velocity estimation and evaluates its effectiveness by comparing it to \emph{Ultimate SLAM} \cite{uslam}, the \gls{sota} monocular \gls{evio} method, as well as by conducting qualitative experiments at highway speeds using full-scale vehicles. We present the proposed method in \Cref{sec:eof_implementation}, and in \Cref{sec:expsetup}, we describe the experimental setup used for the qualitative and quantitative evaluation of our method against \gls{evio} as well as for the full-scale highway experiments.

\subsection{Event Optical Flow Algorithm} \label{sec:eof_implementation}

\subsubsection{Data Preprocessing}
To make use of classical \gls{of} methods, the event data must first be converted into an image-like format. Recently, research has focused on reconstructing greyscale videos from raw events \cite{dvs_survey}. Still, in this work, it has been found that a simple accumulation of events into a 2D histogram is sufficient. Hereby, events are gathered over an accumulation time from which a 2D histogram of triggered events is created. The image on the right in Figure \ref{fig:rgbvsevent} shows an example of such an event histogram. The advantage of this representation is that the accumulation time, analogous to the exposure time in an RGB camera, can be increased or decreased to ensure sharp images at varying vehicle speeds, but the event camera's extremely high capture rate allows for much lower accumulation times, resulting in sharp images, even at very high speeds. Ideally, the accumulation time is set sufficiently high to generate enough features, while not suffering from excesive motion blurr. Given the analysis in \Cref{fig:g0}, the accumulation time needs to be selected, such that the amount of motion blur caused can still be handled by the \gls{of} method. \rev{An alternative is to use histograms with a fixed number of events, which automatically adjusts accumulation time based on velocity. However, this lacks guarantees on maximum blur, so fixed-time accumulation is used instead.}

\subsubsection{Optical Flow Computation}\label{subsec:of_compute}
Using the aforementioned event accumulation, two successive event frames are created and subsequently used to compute \gls{of}. For this purpose, an off-the-shelf implementation of the Farneback \cite{farneback2003two} algorithm from \emph{OpenCV} is used that computes dense, per-pixel optical flow.

\subsubsection{Rigid Body Motion Estimation}\label{subsec:ME}
Given that the camera is mounted facing the ground and no other objects are visible in the frame, the assumption is made that what the camera sees can be considered a single, rigid body moving in 2D. This allows one to simplify the task of finding the vehicle velocity to a case of planar kinematics, where each rigid body's motion can be described by a single translation and rotation. With this interpretation, the calculated \gls{of} vectors can be considered a collection of velocities on the same rigid body. Thus, to describe the vehicle's motion, one must find a rotation $R$ and translation $t$ such that:
\begin{equation}
    \left(R, t\right) = \argmin_{R \in SO(d), t \in \mathbb{R}^d} \sum_{i=1}^{n} \| (R\mathbf{p}_i + t) - \mathbf{q}_i \|^2,
    \label{eq:motion_estimation}
\end{equation}
where $d=2$, $n$ is the number of points for which \gls{of} was calculated, $\mathbf{p_i}$ and $\mathbf{q_i}$ are the initial and final points of the \gls{of} vectors. Plugging in the values from the \gls{of} calculations, this equation can be solved by singular value decomposition using the method of Sorkine-Hornung et al. \cite{motion_estimation_svd}. The resulting translation vector is in pixel space and must be converted to metric space using the camera's focal length and height from the ground. Finally, we use the time delta between the two consecutive frames to convert the translation $t$ of the camera center to a velocity $v_c$ in \SI{}{\metre\per\second} and the rotation $R$ into an angular velocity $\omega$ in \SI{}{\radian\per\second}. We assumed the camera height to be constant, \rev{yet} one could incorporate real-time height measurements (e.g., from a commercial \rev{\gls{tof}} range finder \rev{or sonar sensors as employed by \emph{PX4Flow} \cite{px4flow}}) to account for fluctuations in the vehicle's height.

\subsubsection{Outlier Removal}
In some cases, it can be that the calculated optical flow contains outliers that significantly impact the motion estimation described in \Cref{subsec:ME}. For such scenarios, the motion estimation is integrated into a \gls{ransac} loop as described in \Cref{alg:ransac}. Given a complete input flow, motion estimates are iteratively computed based on a minimum sample of two randomly selected flow vectors. Inliers are determined by comparing the original flow to a vector field of reconstructed flow computed by plugging the current motion estimate into \Cref{eq:transform_vel}. The final motion estimation is calculated using the largest set of inliers found in the \gls{ransac} loop.
\begin{algorithm}[ht]
\caption{RANSAC-based Rigid Body Motion Estimation}
\textbf{Input:} flow $F$, number of iterations $N$, inlier threshold $\varepsilon$\\
\begin{algorithmic}[1]
\STATE $max\_inliers \leftarrow 0$
\STATE $best\_inliers \leftarrow \text{None}$
\FOR{$i = 0$ to $N$}
    \STATE $min\_sample \leftarrow pick\_2\_random\_flows(F)$
    \STATE $(t, r) \leftarrow motion\_estimation(min\_sample)$
    \STATE $vec\_field \leftarrow create\_vector\_field(r, t)$
    \STATE $epe \leftarrow \text{norm}(flow - vec\_field)$
    \STATE $inliers \leftarrow epe < \varepsilon$
    
    \IF{$\text{count}(inliers) > max\_inliers$}
        \STATE $max\_inliers \leftarrow n\_inliers$
        \STATE $best\_inliers \leftarrow inliers$
    \ENDIF
\ENDFOR
\STATE $(t_{final}, r_{final}) \leftarrow motion\_estimation(best\_inliers)$
\end{algorithmic}
\textbf{Output:} Estimated motion $(t_{final}, r_{final})$
\label{alg:ransac}
\end{algorithm}

\subsubsection{Rigid Body Transformation}
The rigid body motion estimation returns the rotation and translation of the camera center with respect to the ground. For integration into the vehicle's state estimation and for comparison with other velocity estimation techniques, the estimated velocity is transformed from the camera center to the rear axle of the car as follows:
\begin{equation}
    \underline{v_A} = \underline{v_c} + \underline{\omega} \times \underline{CA},
    \label{eq:transform_vel}
\end{equation}
where $\underline{v_c}$ and $\underline{\omega}$ are the estimated velocity and yaw rate at the camera center, $\underline{v_A}$ is the velocity at the vehicle's rear axle, and $\underline{CA}$ is the vector from the camera center to the rear axle.

\begin{table*}[ht]
\centering
\vspace{5pt}
\resizebox{0.93\textwidth}{!}
{%
\begin{tabular}{l|ccc|cc|cc}
\toprule
\toprule
\multirow{2}{*}{Method} & \multicolumn{3}{c|}{\bm{$v_{long}$}} & \multicolumn{2}{c|}{\bm{$v_{lat}$}}    & \multicolumn{2}{c}{\bm{$v_{\psi}$}} \\
                        & $RMSE\;[m/s]\downarrow$ & $\sigma\;[m/s]\downarrow$ & $E\;[\%]\downarrow$ & $RMSE\;[m/s]\downarrow$ & $\sigma\;[m/s]\downarrow$  & $RMSE\;[rad/s]\downarrow$ & $\sigma\;[rad/s]\downarrow$ \\
\midrule
\midrule
eVIO \cite{uslam} forward            & 0.0503          & 0.0344          & 2.63          & 0.0466          & 0.0305          & \textbf{0.0783} & \textbf{0.0565} \\ 
eVIO \cite{uslam} downward           & 0.0484          & \textbf{0.0288} & 2.82          & 0.0817          & 0.0552          & 0.2118          & 0.1499 \\ 
eOF \textbf{(ours)}                    & \textbf{0.0470} & 0.0307          & \textbf{2.50} & 0.0487          & 0.0301          & 0.1878          & 0.1172 \\
eOF + IMU \textbf{(ours)}               & \textbf{0.0470} & 0.0307          & \textbf{2.50} & \textbf{0.0287} & \textbf{0.0187} & 0.0969          & 0.0662 \\
\bottomrule
\bottomrule
\end{tabular}
}
\caption{Quantitative evaluation of the velocity states $v_{lon}, v_{lat}, v_{\psi}$ computed by either \gls{evio} (\emph{Ultimate SLAM} \cite{uslam}) or \gls{eof} with respect to motion-capture ground-truth data. All downward-facing results are derived from the same sensor readings. The metrics of each velocity state are reported in terms of \gls{rmse} and its resulting standard deviation $\sigma$. To allow comparison with other experiments conducted at different driving speeds, the relative error ($E[\%]$) is reported for the longitudinal velocity. Because the values for lateral velocity and yaw rate can be zero, a relative error can not be computed for these metrics.}
\label{tab:quant_res}
\end{table*}

\subsubsection{Augmentation with IMU Data}
The contribution of the yaw rate in \Cref{eq:transform_vel} increases proportionally with the distance of the camera center to the rear axle, highlighting the importance of accurate yaw rate estimation. Given the ubiquity of \gls{imu} sensors in robotic platforms, it is a natural choice to supplement the yaw rate estimation with measurements from an \gls{imu} sensor. In this work, we study the effect of replacing the yaw rate estimates from \gls{eof} in \Cref{eq:transform_vel} with raw yaw rate estimates from an \gls{imu}.

\subsection{Experimental Setup} \label{sec:expsetup}
We evaluate the velocity estimation performance using a 1:10 scale open-source autonomous racing platform \cite{forzaeth}. A \texttt{DAVIS 346} event camera, providing \texttt{346x260} pixel resolution for both events (\SI{120}{\dB} dynamic range) and grayscale images (\SI{56.7}{\dB} dynamic range), was mounted on the platform. This camera also includes a 6-axis \gls{imu}. \rev{Compared to newer cameras such as the \texttt{Prophesee EVK4} the \texttt{DAVIS 346} produces noisier events and less clean event frames. Under these circumstances, it is hard to choose good noise thresholds for \gls{ransac}, thus we do not run our method through a \gls{ransac} loop in all experiments using the \texttt{DAVIS 346} camera.} \rev{For the evaluation of the \gls{evio} method, we used \emph{Ultimate SLAM} in its combined mode (combining events, conventional frames, and IMU data) \cite{uslam}. The camera was mounted in two configurations: downward-facing at \SI{0.3}{\meter} above the ground behind the car and forward-facing on top of the center of the car at \SI{0.1}{\meter} as depicted in \Cref{fig:vicon}. For the evaluation of the \gls{eof} method, only the downward-facing configuration was used.}

\subsubsection{Quantitative Measurements}
\begin{figure}[t]
    \centering
    \includegraphics[width=0.9\columnwidth]{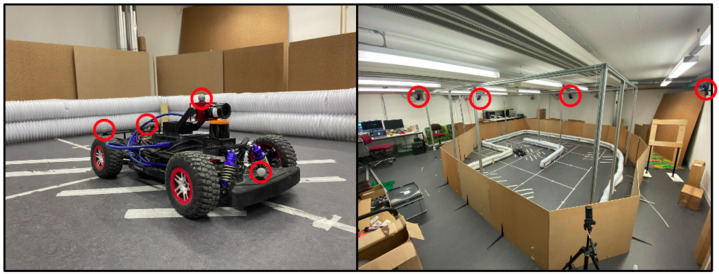}
    \caption{A depiction of the velocity estimation performance in-field experiment. On the left, a 1:10 scaled autonomous racing car is depicted with the \texttt{DAVIS 346} event camera mounted. On the right, the track is set within a motion-capturing setup. The red circles highlight the motion-capture tracking components, i.e., \gls{ir} reflective markers and cameras.}
    \label{fig:vicon}
\end{figure}

\begin{figure*}[htb]
    \centering
    \includegraphics[width=0.9\textwidth]{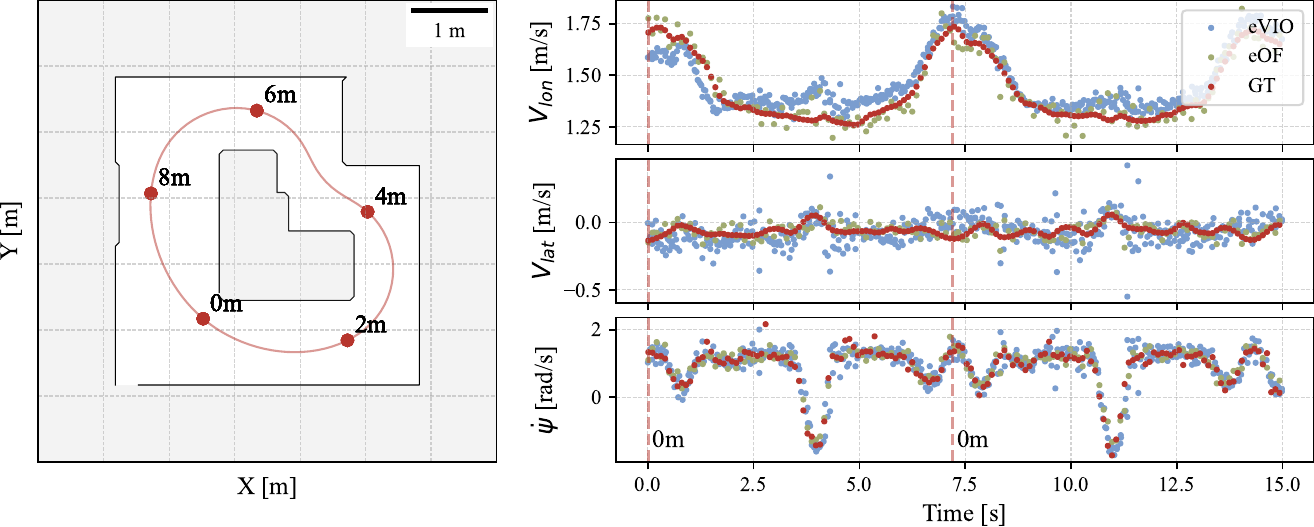}
    \caption{Qualitative visualization of the scaled autonomous driving velocity experiment with motion-capture ground truth comparison. On the left, the tracked trajectory is depicted in red with track advancement $s$ denoted every \SI{2}{\metre}. On the right the $v_{lon}, v_{lat}, v_{\psi}$ velocity states of front-facing \gls{evio} (blue), downwards-facing \gls{eof} (green), and motion-capture ground truth measurements (red). The vertical red dashed lines represent the \SI{0}{\metre} mark of the track depicted on the left.}
    \label{fig:quali_res}
\end{figure*}

\rev{For the collection of the evaluation data, the 1:10 scale car completed ten laps following a given trajectory and velocity profile as depicted in \Cref{fig:quali_res}. Based on the average movement velocity of \SI{1.5}{\meter \per \second}, the accumulation time was set to \SI{33}{\milli\second}. During the accumulation, positive and negative events are collected in separate histograms.} For ground-truth data, a high-precision \si{\milli\metre}-accuracy \emph{Vicon} motion-capture system with six \gls{ir} cameras and reflective markers is used, as shown in \cref{fig:vicon}. The \emph{Vicon} system outputs the marker's pose at \SI{100}{Hz} which are differentiated to obtain velocity and yaw rate measurements. The system's measurable space is limited to \texttt{4x4} \si{\meter}, restricting the maximum velocity. Despite this limitation, the 1:10 scale setup allows accurate evaluations of velocity estimation algorithms. Data acquisition is facilitated through the \gls{ros} integration of the motion-capture system, event camera, and racing car. The results of the quantitative evaluation are later elaborated in \Cref{subsec:results}.

\subsubsection{Qualitative Measurements}
To qualitatively assess the performance of the velocity estimators, two experimental setups were utilized:

\begin{enumerate}[I]
    \item The first setup involved the use of the 1:10 scale car within the motion-capture environment depicted in \cref{fig:vicon}. This setup allows for a visual comparison of how the velocities estimated by both \gls{eof} and \gls{evio} align with the ground-truth measurements, effectively serving as a qualitative subset of the aforementioned quantitative evaluations. These results are further elaborated in \Cref{subsec:quali_vicon_results}.
    \item The second experiment featured a \texttt{Prophesee EVK4} event camera mounted on a full-scale car at \SI{0.635}{\metre} above the ground \rev{using suction cups}, as shown in \cref{fig:title-image}. To avoid saturating the camera's bandwidth from the extremely high event rates generated at highway speeds, only a $50 \times \SI{180}{pixel}$ crop was recorded. The car was driven on a public highway at a constant speed of \SI{32}{\metre\per\second}, measured by \gls{gps}. Subsequently, the raw events were converted to accumulated event histograms with an accumulation time of \SI{100}{\micro\second}.
\end{enumerate}

\section{Experimental Results}
This section details the experimental results in terms of velocity estimation performance of both the \gls{evio} \cite{uslam} and \gls{eof} algorithms. \Cref{subsec:results} presents the quantitative evaluation of the in-field velocity estimation performance between \gls{evio} \cite{uslam} and \gls{eof}. \Cref{subsec:quali_results} qualitatively showcases the velocity estimations alongside the ground-truth motion-capture measurements as well as a full-scale experiment conducted on a real car driving at highway speeds.

\subsection{Quantitative Performance Evaluation}\label{subsec:results}
In \Cref{tab:quant_res}, the quantitative evaluations of the velocity states $v_{lon}$, $v_{lat}$, and $v_{\psi}$ are demonstrated with respect to ground-truth motion-capturing data, as described in \Cref{sec:expsetup}. It is evident that \gls{eof} with \gls{imu} correction achieves the lowest \gls{rmse} values for $v_{lon}$ and $v_{lat}$, with \SI{0.047}{\metre\per\second} and \SI{0.0287}{\metre\per\second}, respectively. \rev{While the improvement in longitudinal velocity estimation is within one standard deviation of the best-performing \gls{evio} configuration, our method achieves a lateral velocity \gls{rmse} that is approximately one standard deviation lower, indicating a notable gain in accuracy.} The results in \Cref{tab:quant_res} also highlight the importance of incorporating \gls{imu} data in the \gls{eof} implementation. The yaw rate error ($v_{\psi}$ \gls{rmse}) of the \gls{imu} is $48.4\%$ smaller than that of the pure \gls{eof} implementation, resulting in a direct improvement of the lateral velocity estimation ($v_{lat}$ \gls{rmse}) of $41.1\%$. 
\rev{It must be noted that the yaw rate error of our method in the \gls{eof} + \gls{imu} configuration is still $19\%$ worse than that of the forward-facing \gls{evio}. This is most likely due to the fact that our implementation uses raw \gls{imu} measurements while the \gls{evio} estimate is filtered through an \gls{ekf}.} Because the camera is mounted perpendicularly behind the rear axle, the yaw rate has no influence on the longitudinal velocity in \Cref{eq:transform_vel}, thus $v_{lon}$ \gls{rmse} is the same for both \gls{eof} implementations. Compared to forward-facing \gls{evio} ($v_{lon}$ \gls{rmse}: \SI{0.0503}{\metre\per\second}, $v_{lat}$ \gls{rmse}: \SI{0.0466}{\metre\per\second}), this represents a 6.5\% improvement in $v_{lon}$ \gls{rmse} and a 38.3\% improvement in $v_{lat}$ \gls{rmse}. \rev{These results serve to indicate that the more direct approach of measuring velocity directly from optical flow instead of deriving it from a pose graph can lead to improved accuracy, especially in the lateral direction.} Conversely, the yaw rate error ($v_{\psi}$ \gls{rmse}) is the lowest with the forward-facing \gls{evio}, resulting in a $v_{\psi}$ \gls{rmse} of \SI{0.0783}{\radian\per\second}. Compared to \gls{eof} with \gls{imu} ($v_{\psi}$ \gls{rmse}: \SI{0.0969}{\radian\per\second}).

\subsection{Qualitative Evaluation}\label{subsec:quali_results}
This section presents the qualitative assessment of the velocity estimators, expanding on the quantitative analysis provided earlier. The qualitative evaluation is conducted through two distinct experiments: one with a scaled model and another in a full-scale, real-world scenario.

\subsubsection{Scaled Velocity Evalution}\label{subsec:quali_vicon_results}
In alignment with the quantitative findings presented in \Cref{subsec:results}, \Cref{fig:quali_res} provides a qualitative comparison of velocity estimation for the 1:10 scale autonomous racing car using \gls{evio} and \gls{eof}. The \gls{evio} results are shown in blue, with the sensor configured in the downward-facing orientation. The \gls{eof} results, depicted in green, are derived from the same set of image data used for the downward-facing \gls{evio}. This consistent setup allows for direct comparison between the two methods, utilizing the same ground truth measurements, which are displayed in red. \rev{Both \gls{evio} and \gls{eof} have been empirically validated for closed-loop control, demonstrating their practical applicability.} Interestingly, the \gls{evio} algorithm tends to overestimate the longitudinal velocity compared to \gls{eof} in the lower-speed sections of the track. It can also be seen that \gls{eof} tracks the lateral velocity much better compared to \gls{evio}, visually confirming the findings shown in \Cref{tab:quant_res}.

\begin{figure}
    \centering
    \includegraphics[width=0.88\linewidth]{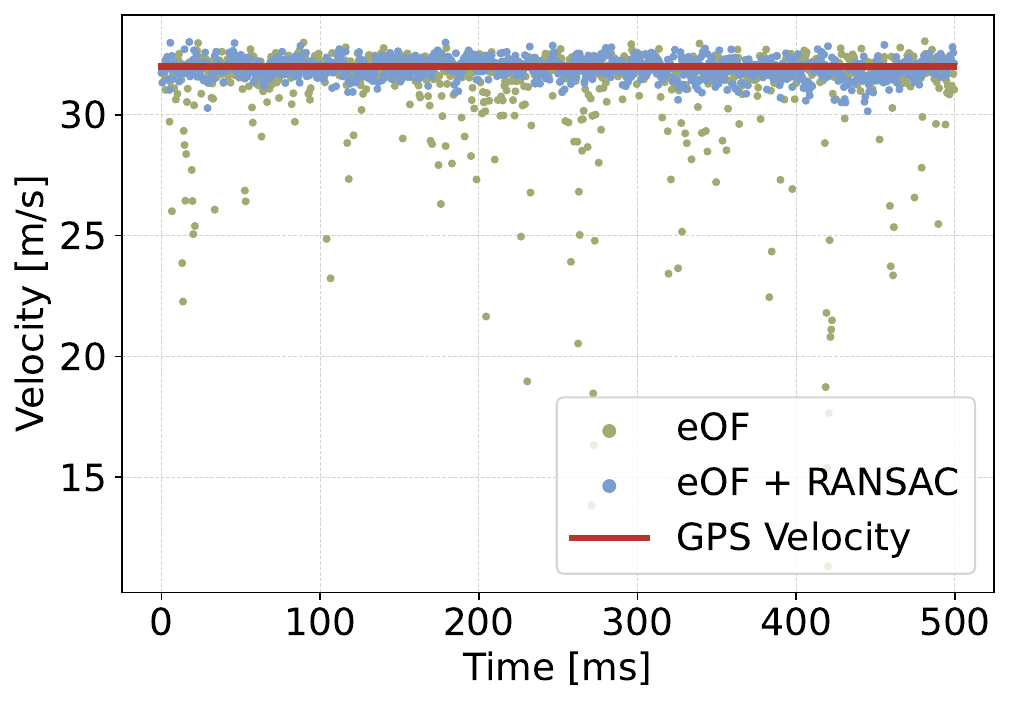}
    \caption{Estimated velocity using the proposed \gls{eof} method, with and without \gls{ransac}, compared to the velocity measured by \gls{gps}.}
    \label{fig:highway_vel}
\end{figure}

\subsubsection{Full-Scale Highway}\label{subsec:quali_clio_results}
The proposed \gls{eof} method is evaluated over the \SI{500}{\milli\second} highway recording, corresponding to $5000$ consecutive frames.
\rev{Since the suction-cup mount could not ensure perfect alignment with the car's longitudinal axis, camera velocity may include lateral components even during straight motion. Thus, only the magnitude of the planar velocity is reported, approximating the car’s longitudinal speed.}
In \Cref{fig:highway_vel} it can be seen that estimating the velocity without \gls{ransac} leads to a significant amount of large outliers. In this case, the mean estimated velocity is \SI{31.28}{\metre\per\second} ($E$ $2.3 \%$) with a standard deviation of \SI{2.03}{\metre\per\second} compared to \SI{31.88}{\metre\per\second} ($E$ $0.4 \%$) with a standard deviation of \SI{0.64}{\metre\per\second} when utilizing \gls{ransac}. \rev{Due to the relatively low resolution of $50 \times \SI{180}{pixels}$, the \gls{eof} algorithm's latency decreases to \SI{2}{\milli\second} without \gls{ransac} and rises to \SI{13}{\milli\second} when \gls{ransac} refinement is applied.} \revdel{Due to the relatively small resolution of 50x180 pixels the latency of the \gls{eof} algorithm drops to 2ms when not using \gls{ransac} and to 13ms when refining estimates with \gls{ransac}.} \rev{An inlier threshold of \SI{0.5}{pixels} and an iteration count $N$ of 16 was used in this implementation of \gls{ransac}.}These results demonstrate that event cameras are capable of capturing meaningful data at very high speeds and that the proposed \gls{eof} method can effectively process this data to produce accurate velocity estimates in real-world driving scenarios. \rev{The proposed approach handled real-world imperfections like cracks and uneven roads. Scale adjustment using a laser distance sensor is advised to address height changes caused by the suspension. Limitations of the proposed approach include featureless or flat surfaces, which prevent optical flow calculation.}

\subsection{Computational Results}
\rev{
The \gls{cpu} load of the \gls{ros} nodes of either \gls{evio} or \gls{eof} was measured using the \texttt{psutil} tool, where 100\% utilization represents full load of a single core. \Cref{tab:compute} presents the computational load, memory, and latency of the algorithms under investigation. The results indicate that both \gls{evio} and \gls{eof} exhibit very similar computational loads, differing by only 6.96 percentage points ($\mu_{\text{cpu},\text{eOF}}$: 98.39\%, $\mu_{\text{cpu},\text{eVIO}}$: 105.35\%). A similar pattern is observed with the memory usage, resulting in a marginal \SI{8.32}{\mega\byte} difference ($\mu_\text{mem,\text{eOF}}$: \SI{118.12}{\mega\byte}, $\mu_\text{mem,\text{eVIO}}$: \SI{126.44}{\mega\byte}). However, when considering latency, \gls{evio} achieves a significantly higher output frequency for odometry solutions, approximately 60 times faster than \gls{eof} ($\mu_{t,\text{eOF}}$: \SI{60.35}{\milli\second}, $\mu_{t,\text{eVIO}}$: \SI{1.04}{\milli\second}). Yet it is important to notice that the \gls{evio} algorithm operates asynchronously with its global loop closure, meaning that the odometry output frequency extrapolates its pose on a fixed time interval. Therefore, the latency of the asynchronous \gls{evio} cannot be directly compared with the synchronous computation of \gls{eof}.}
 
\begin{table}[t] 
    \vspace{2mm}
    \centering 
    \begin{tabular}{l|cc|cc|cc}
    \toprule
    \textbf{Method} & \multicolumn{2}{c|}{\textbf{\gls{cpu} [\%]$\downarrow$}} & \multicolumn{2}{c|}{\textbf{Mem [MB]}$\downarrow$} & \multicolumn{2}{c}{\textbf{Latency [ms]}$\downarrow$} \\
    & $\mu_{cpu}$ & $\sigma_{cpu}$ & $\mu_{mem}$ & $\sigma_{mem}$ & $\mu_{t}$ & $\sigma_{t}$ \\
    \midrule
    \gls{evio} & 105.35 & 16.98 & \textbf{118.12} & 4.17 & \textbf{1.04}$\dagger$ & \textbf{5.85}$\dagger$ \\
    \gls{eof} & \textbf{98.39} & \textbf{3.92} & 126.44 & \textbf{2.17} & 60.35 & 10.40 \\
    \bottomrule
    \end{tabular}%
    \caption{\rev{Computational results of \gls{evio} and \gls{eof}. \gls{cpu} load and memory usage using the \texttt{psutil} tool. The \gls{cpu} load, memory usage, and latencies are reported with their mean $\mu$ and standard deviation $\sigma$ and are all computed on an \texttt{i5-10210U} \gls{cpu}. $\dagger$ denotes that the algorithm is asynchronous, meaning that latency can not clearly be compared.}}
    \label{tab:compute}
\end{table}

\section{Conclusion}
In this work, we introduce an accurate \gls{eof}-based method for direct velocity measurement from novel event camera data without needing to construct a pose graph. Like \gls{vio}, this approach decouples the velocity estimation from unreliable proprioceptive sensors, such as wheel encoders, thus avoiding the need for complex motion models and tire dynamics. Our quantitative experiments show that not only does the simple proposed method yield results on par with \gls{evio}, \rev{but the direct \gls{eof} approach can even outperform the more algorithmically complex \gls{sota} \gls{evio} method, especially in lateral velocity estimation where the accuracy is improved by $38.3\%$.} \revdel{but the direct \gls{eof} approach can even outperform the more algorithmically complex \gls{sota} \gls{evio} method in both longitudinal and lateral velocity estimation by $6.5\%$ and $38.3\%$, respectively.} Additionally, qualitative experiments at highway speeds, up to \SI{32}{\metre\per\second}, highlight the method's strong potential for real-world deployment. \rev{In future work, one could train a dedicated optical flow network on downward-facing event data to explore different event representations and potentially achieve better flow estimates. Further, our \gls{eof} approach could be validated on additional real-world data.}







\FloatBarrier
\bibliographystyle{IEEEtran}
\bibliography{main}

\end{document}